\documentclass[preprint,12pt]{elsarticle}

\usepackage{algpseudocode}
\usepackage{amsmath}
\usepackage{amssymb}
\usepackage[english]{babel}
\usepackage{blindtext}
\usepackage{booktabs}
\usepackage{caption}
\usepackage{geometry}
\usepackage{graphicx} 
\usepackage{mhchem}
\usepackage{natbib}
\usepackage{soul}
\usepackage{subcaption}
\usepackage{tikz}
\usepackage{xcolor}

\journal{journal}

\begin{document}

\begin{frontmatter}



\title{Fast Uncertainty Quantification of Spent Nuclear Fuel with Neural Networks}

\author[psi,eth]{Arnau Alb\`a}
\author[psi]{Andreas Adelmann}
\author[psi,eth]{Lucas M\"unster}
\author[psi]{Dimitri Rochman}
\author[psi]{Romana Boiger\corref{corr_auth}} 
\cortext[corr_auth]{Corresponding author: Romana Boiger, romana.boiger@psi.ch}

\affiliation[psi]{organization={Paul Scherrer Institut},
            addressline={PSI, Forschungstrasse 111},
            city={Villigen},
            postcode={5232},
            country={Switzerland}}

\affiliation[eth]{organization={ETH Z\"urich},
            addressline={R\"amistrasse 101},
            city={Zurich},
            postcode={8092},
            country={Switzerland}}

\begin{abstract}
The accurate calculation and uncertainty quantification of the characteristics of spent nuclear fuel (SNF) play a crucial role in ensuring the safety, efficiency, and sustainability of nuclear energy production, waste management, and nuclear safeguards. State of the art physics-based models, while reliable, are computationally intensive and time-consuming. This paper presents a surrogate modeling approach using neural networks (NN) to predict a number of SNF characteristics with reduced computational costs compared to physics-based models. An NN is trained using data generated from CASMO5 lattice calculations. The trained NN accurately predicts decay heat and nuclide concentrations of SNF, as a function of key input parameters, such as enrichment, burnup, cooling time between cycles, mean boron concentration and fuel temperature. The model is validated against physics-based decay heat simulations and measurements of different uranium oxide fuel assemblies from two different pressurized water reactors.
In addition, the NN is used to perform sensitivity analysis and uncertainty quantification. The results are in very good alignment to CASMO5, while the computational costs (taking into account the costs of generating training samples) are reduced by a factor of 10 or more.
Our findings demonstrate the feasibility of using NNs as surrogate models for fast characterization of SNF, providing a promising avenue for improving computational efficiency in assessing nuclear fuel behavior and associated risks. 
\end{abstract}






\nonumnote{\textit{Abbreviations:} decay heat (DH), end of life (EOL), fuel assembly (FA), neural network (NN), pressurized water reactor (PWR), uncertainty quantification (UQ), sensitivity analysis (SA).}

\end{frontmatter}


\section{Introduction}


The total amount of spent nuclear fuel worldwide sums up to 392 000 tons of heavy metal in storage and reprocessing \cite{international2022iaea}. Given the high levels of radioactivity, it is crucial to focus on safety considerations regarding fuel handling, transportation, storage, and disposal. Therefore, it is important to understand the characteristics of spent nuclear fuel (SNF) and accurately predict their uncertainties, because they can provide valuable information about the behavior of the fuel and the potential risks. To ensure the safe storage and disposal of SNF, characteristics like nuclide concentrations and decay heat (DH) must be predicted in advance for the next thousands of years.

The topic of SNF characterization from the neutronics aspect was discussed in \cite{Rochman2023}, comparing measured nuclide concentrations and decay heat to simulation results. In addition the paper  provides recommendations for best practices, biases, uncertainties, and improvement of code prediction capabilities for SNF modeling.

Since measurements of nuclide concentrations and decay heat over a very long time period are obviously not possible, these quantities are usually predicted by means of computational models. Examples of such models are stated in \cite{Rochman2023}, e.g. CASMO5, EVOLCODE, SCALE/Polaris, ALEPH, MURE or TRITON. Within this paper specifically, the physics-based modeling software CASMO5 \cite{Casmo5} was used, an advanced lattice physics code designed for the modeling of pressurized water reactor (PWR) and boiling water reactor fuel. It is specifically optimized to handle complex fuel design, including high mixed-oxide concentrations and high burnable poison concentrations.

Several studies have used such computational models in the past to investigate the uncertainty of SNF, e.g. in \cite{WOS:000974744000001,shama_validation_2022} the biases, uncertainties, and correlations of calculated decay heat from SNF using the Polaris and ORIGEN codes were analyzed, finding that both codes exhibited insignificant biases and similar uncertainties and correlations influenced by fuel assembly (FA) burnup and cooling time. The study also made use of machine learning models and the MOCABA algorithm for predicting biases and verifying results. In \cite{WOS:000678338400004} a two-step analysis to quantify the uncertainty and sensitivity of decay heat and nuclide densities in a depleted light-water FA was conducted. It was found that the cooling time affects the decay heat uncertainty, while nuclear data and assembly design parameters contribute the most to the uncertainty, with a few nuclides playing a significant role. Uncertainty sources from technological uncertainties, modeling assumptions, modeling resolution and nuclear data uncertainties on PWR nuclide inventories for severe accidents were analysed in \cite{Ichou2023}. They found that the nuclear data and the "infinite lattice approximation" have the biggest influence on the bias and uncertainties. To reduce the latter one, in  \cite{Seidl2023} the potential of making the prediction of decay heat and other source terms for spent nuclear fuel more accurate was considered.

The major drawback of such physics-based models is that they are computationally intensive, CASMO5 needs minutes or even hours for one forward simulation for one single FA, depending on the details of the irradiation history (note that CASMO5 is one of the fastest available physics-based codes).  Furthermore, if the calculations are to be accompanied with uncertainties and sensitivity analysis, the number of required simulations for a single FA can increase by more than 100. So in order to be able to predict uncertainties of SNF characteristics for thousands of FAs for the next thousands of years, there is a high demand for faster methods for such calculations. 

Therefore, machine learning approaches, like surrogate models to replace time-consuming, complex calculations, offer new possibilities in various disciplines. A popular ansatz to construct surrogate models is the use of neural networks (NN). They can approximate any function according to the universal approximation theorem \cite{HORNIK1989359}.
An overview of the applications of machine learning methods, like neural networks, specifically for the disposal of high-level nuclear waste, is given in \cite{HU2023109452}, whereas in \cite{Nissan2019}, a review of artificial intelligence methods for in-core fuel management was conducted.

In \cite{Bae2020} a neural network model to predict the composition of PWR SNF based on initial enrichment and burnup was trained with data coming from the SNF Storage, Transportation \& Disposal Analysis Resource and Data System Unified Database. The trained model was validated with the U.S. SNF inventory profile and showed errors of less than 2\% and fast computation time of 0.27s for 100 predictions. Similar models, predicting nuclide concentration based on enrichment and burnup, were used in \cite{lei_prediction_2021, Lei2022}, where the DRAGON code was used to generate training samples. In that case the constructed models were linear, random forest, plain NNs, and NNs with dropout regularisation. It was found that the latter method outperformed the rest, although benchmarking was done only with the DRAGON simulations.


In \cite{ebiwonjumi_machine_2021}, surrogate models of three types, namely NNs, Gaussian processes, and support vector machines, were trained on experimental measurements and used to predict the decay heat and perform an uncertainty quantification. In that case however, the quality of the surrogates was limited by the small amount of experimental data available. As a remedy, it was attempted to increase the amount of data by generating so-called "synthetic data", based on the same experimental data. However, results showed no significant improvement when using synthetic data versus the experimental data.

The approach in the present work was similar to \cite{ebiwonjumi_machine_2021}, and even used some of the same data for benchmarking the NN, but differed in a crucial aspect: instead of using scarce experimental data, or synthetic data that might not make physical sense, we used reliable physics-based simulations with CASMO5, to create a dataset for training the NN. The advantage of this approach is that we were able to generate a larger and more representative training dataset, without the noise from experimental measurements. The generated dataset included only one specific FA from a PWR, with randomly sampled characteristics.
The trained neural network was then a surrogate model for characterising SNF, which could be described as the following map:

\begin{equation*}
f:  \underbrace{\begin{pmatrix}
            \texttt{Enrichment} \\
            \texttt{Burnup} \\
            \texttt{Cooling time between cycles}\\
            \texttt{Mean boron concentration}\\
            \texttt{Fuel temperature}\\
        \end{pmatrix}}_{\text{Input: Fresh fuel characteristics and irradiation history}}
        \rightarrow
        \underbrace{
        \begin{pmatrix}
            \texttt{Decay heat } (t) \\
            \texttt{Actinide concentration }\\
            \texttt{Concentration of \ce{^{137}Cs} and \ce{^{90}Sr}} \\
        \end{pmatrix}}_{\text{Output: SNF characteristics}}\,,
\end{equation*}

with time $ t\in[2,1000]$ years, and the nuclide concentrations being predicted at the end of life (EOL) of the FA.

The previous works \cite{Bae2020,lei_prediction_2021,Lei2022} trained similar models from physics-based simulations, although they only had two input parameters, namely the enrichment and the burnup. Here we found that including more input parameters played an important role, especially in the prediction of certain nuclides, as will be shown in the sensitivity analysis of section \ref{sec:UQandSA}. Further novelties in our work are the comparison of the NN predictions to experimental results, and the application of the NN for uncertainty and sensitivity analysis. The latter application is especially promising, since it is a case where a large number of simulations are required, and the NN as a cheap model, in terms of computation time (one forward simulation lasts $5\cdot10^{-4}$ s), could lead to substantial improvements in computational costs. The results of uncertainty quantification and sensitivity analysis were validated against CASMO5 to ensure accuracy and reliability. 

The goal of this paper is to show the feasibility of using surrogate models, specifically neural networks, trained on physics-based simulations to predict SNF characteristics with reduced computational cost but the same accuracy as physics-based models. 

The paper is organized in the following way. First, the methodology, including the experimental data, physics-based data, as well as neural networks, sensitivity analysis and uncertainty quantification are explained in detail. 
In the results section, the performance of the NN is presented. Different hyperparameters and dataset sizes for training the NN are tested and compared to each other. The NN predictions are compared to the available measurements and to physics-based simulations.  Finally, sensitivity analysis and uncertainty quantification carried out with the NN and CASMO5 are compared in terms of accuracy and computational time. Concluding remarks finish the paper.

\section{Material and Methods}
\subsection{Neural Network}

Neural networks are machine learning models inspired by the structure and function of the human brain. They consist of interconnected neurons, organized in layers, that process and transform data through weighted connections and biases and activation functions. By leveraging these connections and using non-linear activation functions,  NN can learn arbitrarily complex patterns in the data, \cite{HORNIK1989359}. For the hidden layers we used the non-linear activation function ReLU, $\text{ReLU}(x) = \max (0,x)\,$. 
The goal of training a NN is to reproduce a given output as accurately as possible, by adjusting the weights and biases. This is done by formulating an optimization problem, where the loss function is minimized w.r.t. weights and biases. The choice of loss function is highly dependent on the problem to model. As we value the accuracy of all 53 outputs equally, we chose a general mean squared error (MSE) as loss function for the regression problem, given by $\text{MSE} = \frac{1}{N}\sum_{i=1}^N (y_i - \hat y_i)^2$ where $y_i$ is the predicted value and $\hat y_i$ is the true value.

As stochastic optimizer we chose the ADAM method \cite{kingma2017adam}. It takes into account the second moment of the gradient which can allow for faster convergence of the optimization problem than with basic stochastic gradient descent.

We implemented the NN with PyTorch, a highly customizable machine learning framework with GPU acceleration capabilities \cite{pytorch}.
To compare the performance of different networks, we split the dataset into training and test datasets, meaning that we separated 200 samples for testing from the beginning. The training dataset was further split into the training and validation datasets, with a ratio of $80\%-20\%$. While we trained the network on the training dataset, we used the validation dataset only to track performance during training. If the loss on the validation dataset does not improve we can abort the training process as the network has converged, so early stopping is utilized. It is common practice to normalize the input and output data to have a mean of 0 and a standard deviation of 1. This reduces the risk of introducing biases from quantities with a large difference in magnitude.

Parameters that describe the architecture and learning process of the network are called hyperparameters. The hyperparameters in this study are the number of hidden layers, number of neurons per layer, batch size, learning rate (used in the optimization algorithm). The choice of hyperparameters depends on the problem and is crucial as it determines the accuracy of a model.

\subsection{CASMO5 Lattice Code}

The training samples for the NN were obtained from physics-based simulations with the neutron transport code CASMO5 \cite{Casmo5}, a deterministic 2D lattice code from the Studsvik Core Management System. This code takes as input the characteristics, geometry, and irradiation history of a fresh FA, and simulates its irradiation history up to its end of life by repeatedly solving the neutron transport equation and the decay (Bateman) equation, with a predictor-corrector scheme. Each simulation provides a plethora of information as output regarding reactor operation and fuel characteristics. Of these outputs, the quantities of interest in this work were the DH at several cooling times and a number of nuclide concentrations at EOL. Section \ref{sec:prep_training_set} contains further details on the outputs of interest.

Previously, in \cite{shama_validation_2022}, CASMO5 was already used for predicting the FAs measured at the Clab facility \cite{sturek_measurements_2006}, in a study where several neutron transport codes were validated against decay heat measurements. In that study it was reported that the CASMO5 simulations overpredicted the DH with an average bias of $+0.9\%$, with respect to experiments (note that in \cite{shama_validation_2022} the employed nuclear data library was ENDF/B-VII.1, and results may differ from the present paper where ENDF/B-VII.0 was used instead). 

Regarding the technical details of the computations, the FAs were assumed to have a 4-fold symmetry (see fig. \ref{fig:FA_sketches}) and hence only one quarter of the geometry was simulated. Reflecting boundary conditions were used, and the guiding tubes were assumed to be filled with water, since no information was available regarding the control rods. The nuclear data library used for CASMO5 was based on ENDF/B-VII.0 \cite{chadwick_endfb-vii0_2006}, and the default 19 discrete energy groups were used. As was done in \cite{sturek_measurements_2006, shama_validation_2022}, all the cooling and burnup cycles were simulated as single steps with cycle-average parameters, i.e. for each cycle CASMO5 performed a single burnup (or depletion) step, where the values of the power, boron concentration, and  fuel temperature were the average of these quantities in the current cycle. Due to this last convention, the simulations of the FAs in the Clab report \cite{sturek_measurements_2006} were performed with a small number of timesteps, between 3 and 8 steps, and were thus relatively computationally cheap. The average computation time for the simulations in this work was $(58\pm 11)$ seconds, and were carried out on single CPU cores, specifically on Intel Xeon Gold 6152 processors at 2.10 GHz with 4 GB of RAM. Nevertheless, it should be noted that commonly CASMO5 calculations use a more detailed burnup history with multiple small burnup steps within each cycle, and in such cases a single simulation requires significant longer computational time.

\subsection{Training Data}
\label{sec:prep_training_set}

The training dataset was generated by executing a large number of CASMO5 simulations with randomly sampled input parameters. The varied parameters were enrichment, burnup, mean fuel temperature, mean boron concentration in the coolant, and total number of cooling days in-between burnup cycles. The remaining input parameters of the simulations were kept constant, and were based on the assembly C20 from the Clab report \cite{sturek_measurements_2006}. The ranges of the uniformly sampled input parameters are shown in table \ref{tab:input_ranges}, which were chosen to include the parameters of all the FAs presented in the Clab report. A detailed explanation of how the CASMO5 input files were written can be found in the \ref{app:generation_input}.

\begin{table}[h!]
    \centering
    \begin{tabular}{|l l|}\hline
         Enrichment [$\%$] & [1.5, 5.5] \\
         Burnup [MWd/kgU] & [5, 70]\\
         Fuel temperature [k] & [750, 950]\\
         Mean boron concentration [ppm] & [100, 1000]\\
         Cooling time between cycles [days] & [50, 3200]\\
         \hline
    \end{tabular}
    \caption{Ranges used for uniform random sampling of the input quantities of the training dataset.}
    \label{tab:input_ranges}
\end{table}

The C20 assembly, which was used as a basis for the training dataset, was a $15\times 15$ UO$_2$ FA (fig. \ref{fig:FA_sketches_R2}) burned in a PWR for 4 burnup cycles. Therefore all the simulations from the training dataset were for this specific type of FA, with only variations in the aforementioned input parameters. Any FAs with a different pin arrangement, number of cycles, fuel density, or other differences were not represented in this dataset. Despite this, as will be shown in the results section \ref{sec:NN_performance}, the resulting NN was successfully applied to a wide variety of FAs, even those outside of the training dataset range.

Regarding the output of the simulations, the quantities of interest that were gathered were the decay heat, the concentration of actinides, and the concentration of the radioactive nuclides $^{90}$Sr and $^{137}$Cs. The nuclide concentrations were only calculated at EOL, whereas the decay heat was calculated at 2,5,10,11,12,...,28,29,30,100,1000 years of cooling after EOL. This choice of cooling time was motivated by the measurements presented in \cite{sturek_measurements_2006}, where all the measured FAs had a cooling time ranging between 12 and 27 years (see tab. \ref{tab:FA_ranges}).

\subsection{Experimental Data}

To validate the trained NN, the decay heat  of several existing FAs was calculated with the NN and neutron transport codes, and compared to experimental measurements performed on these FAs. For this validation study, 34 depleted uranium oxide (UO$_2$) FAs were considered, from Swedish PWRs. These FAs were part of several measurement campaigns in 2003 and 2004, the results of which can be found in a Clab report from 2006 \cite{sturek_measurements_2006}. In total, 43 measurements were carried out for the 34 studied FAs from PWRs, with some assemblies being measured multiple times on different dates.

Additionally, \cite{sturek_measurements_2006} also contains the detailed burnup history, cooling cycles, and initial characteristics of the FAs (pages 264-266 and 271-273). In the report, the FA data was used to produce computational models of each FA with the SCALE code system \cite{SCALE2018} and predict the DH, which could then be validated against experiments. It was reported that the SCALE calculations overpredicted the DH with an average bias of $+2\%$ with respect to the measurements.

\begin{figure}[h!]
    \centering
    \begin{subfigure}[b]{0.4\textwidth}
        \centering
        \includegraphics[width=\textwidth]{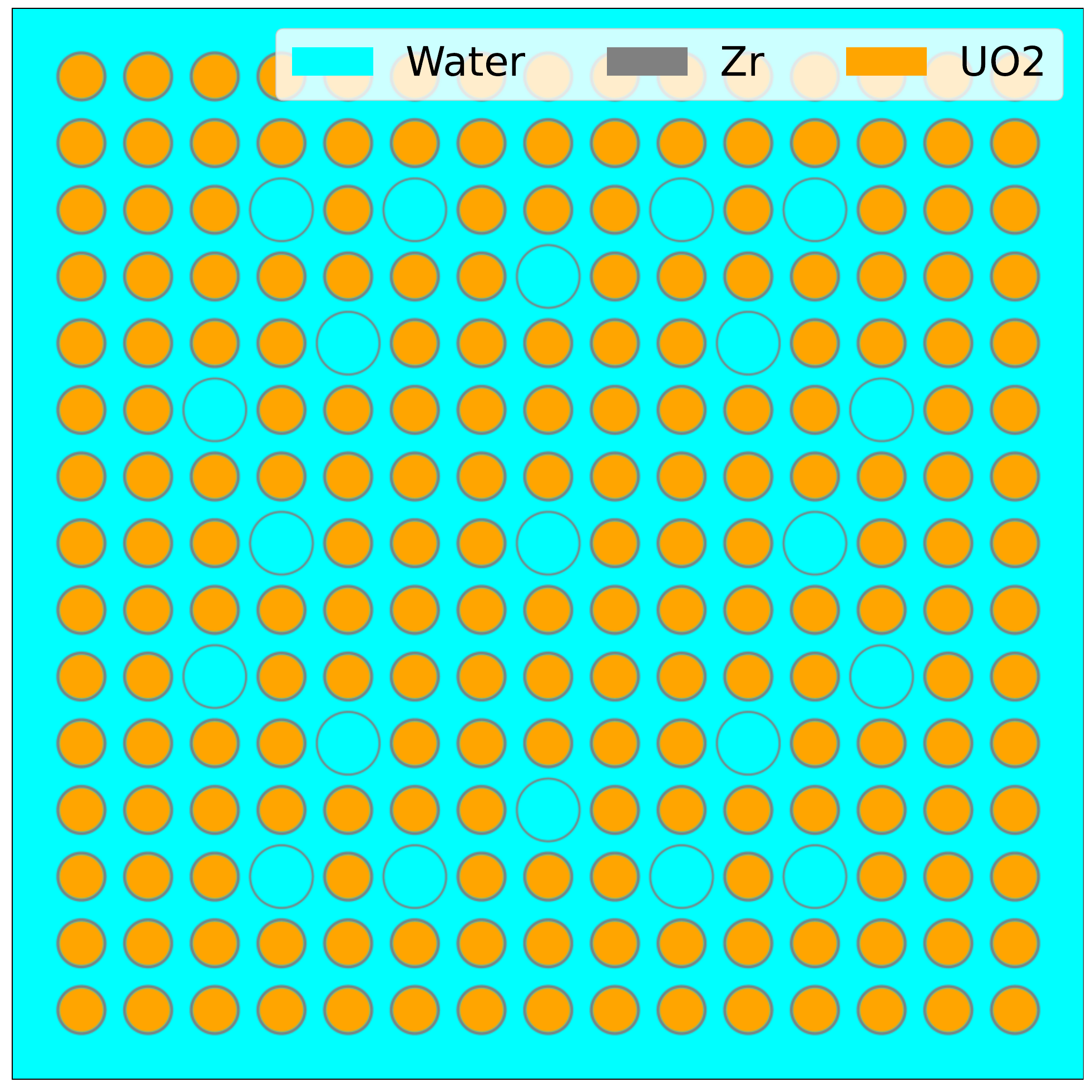}
        \caption{Ringhals-2 assembly with a $15\times 15$ arrangement.}
        \label{fig:FA_sketches_R2}
    \end{subfigure}
    \hspace{5mm}
    \begin{subfigure}[b]{0.4\textwidth}
        \centering
        \includegraphics[width=\textwidth]{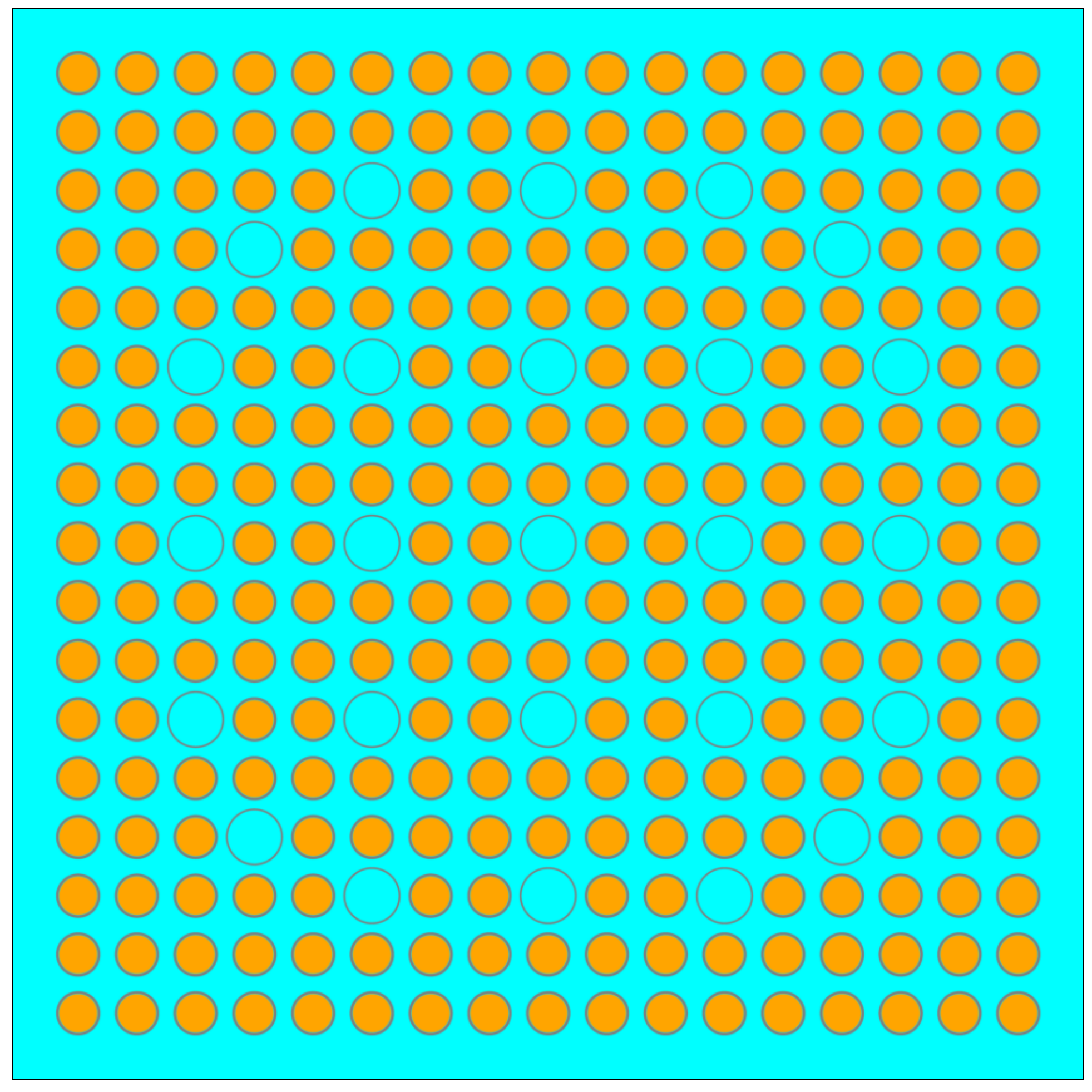}
        \caption{Ringhals-3 assembly with a $17\times17$ arrangement.}
    \end{subfigure}
    \caption{Slices of the FAs from the Clab report \cite{sturek_measurements_2006}. The guiding tubes are simulated as being filled with water.}
    \label{fig:FA_sketches}
\end{figure}

The studied FAs were of two types: 23 came from the Ringhals-2 PWR, with a $15\times15$ arrangement of 204 UO$_2$ pins with zirconium alloy (Zr) cladding and 21 empty Zr-guiding tubes, and the other 20 came from the Ringhals-3 PWR with a $17\times17$ arrangement of 264 fuel pins and 25 guiding tubes. Figure \ref{fig:FA_sketches} shows a 2D slice of each FA type. Table \ref{tab:FA_ranges} shows the ranges of the reported FA characteristics.

\begin{table}[h!]
    \centering
    \begin{tabular}{|l l|}\hline
         Enrichment [$\%$] & [2.1, 3.4] \\
         Burnup [MWd/kgU] & [19.7, 51.0]\\
         Fuel temperature [k] & [750, 950]\\
         Mean boron concentration [ppm] & [287.0, 430.8]\\
         Fuel density [g/cm$^3$] & [10.0, 10.4]\\
         Pellet diameter [mm] & [8.2, 9.3]\\
         Rod diameter [mm] & [9.5, 10.75]\\
         Cladding thickness [mm] & [0.57, 0.73]\\
         Cooling time between cycles [days] & [96, 3128]\\
         Cooling time when measured [years] & [12.9, 23.2]\\
         Measured decay heat [W/t] & [498.8, 1583.6]\\
         \hline
    \end{tabular}
    \caption{Range of characteristics of the FAs measured in \cite{sturek_measurements_2006}.}
    \label{tab:FA_ranges}
\end{table}

\subsection{Uncertainty Quantification and Sensitivity Analysis}

The goal of uncertainty quantification (UQ) is to study the output quantities of simulations or experiments with undetermined inputs. In the context of SNF, these uncertainties around input parameters are related to many factors, such as the uncertainty in parameters of importance during the irradiation history and limited measurement precision.

Sensitivity analysis (SA) aims to find the contribution of each input parameter for every output. A detailed understanding of uncertainties and the importance of all input parameters makes it possible to improve performance by reducing the uncertainty of specific parameters. 

We performed uncertainty quantification and sensitivity analysis with Monte Carlo (MC) methods. Specifically, we considered a multivariate normal distribution with 5\% standard deviation around all relevant input parameters of two fuel assemblies C01 and C20 from the Clab report \cite{sturek_measurements_2006}. The accuracy of this approach is dependent on the number of samples $N$ used. As MC methods converge as $\frac{1}{\sqrt{N}}$, increasing the number of samples is computationally inefficient when running expensive CASMO5 calculations. An NN  with fast evaluation times allows us to perform these methods with a significantly larger number of samples. Note, that an uncertainty of 5\% on every quantity is not necessarily physical and was chosen arbitrarily to demonstrate the capabilities of neural networks.

To validate the UQ done with the NN, the modeled mean and standard deviation were directly compared to CASMO5 simulations over 1000 samples. For SA we generated 1536 samples for every FA and used the Sobol' method \cite{sobol_sensitivity_1993} provided by the open source library SALib \cite{Iwanaga2022, Herman2017}. Sobol' analysis is a variance-based method that assumes the output variance can be attributed to fractions of the variance of the different input quantities. The Sobol' method allows the calculation of first-order indices that indicate the direct contribution of variance from one input to one output. Additionally, the total-order indices can be calculated, which represent the complete influence of input to output variance taking into account higher order interactions between input parameters.

\section{Results and Discussion}
\subsection{Performance of the Neural Network}
\label{sec:NN_performance}

The performance of the neural networks was evaluated through comparison with measurements of FAs from the Clab report \cite{sturek_measurements_2006} as well as calculations based on the same FAs by CASMO5.

The choice of hyperparameters of the NN is considered as an optimization problem to find the model with the highest prediction accuracy. We define the space of possible hyperparameters in table \ref{tab:hyperparameter space} which we explore with a Tree-structured Parzen Estimator as provided by the open source python library Optuna \cite{optuna_2019}. This method uses Bayesian optimization in an attempt to find the hyperparameters that minimize the average MSE on the validation dataset over the last 1000 batch iterations, where a batch iteration describes one evaluation of the network and consequent weight adjustment. A training epoch is complete after the entire training set has been evaluated once.

\begin{table}[h!]
\centering
\begin{tabular}{|ll|}
\hline
Hyperparameter & Space \\
Hidden layers  &  $\{l\in\mathbb N\mid1\leq l\leq 5\}$     \\
Neurons        &  $\{d\in\mathbb N\mid50\leq d\leq 1000\}$     \\
Learning rate  &  $[0.0001, 0.005]$    \\
Batchsize      &  $\{8, 16, 32, 64, 128\}$     \\
Epochs         &  1000     \\ \hline
\end{tabular}
\caption{Hyperparameter domain to explore. The learning rate is sampled in log steps.}
\label{tab:hyperparameter space}
\end{table}

Table \ref{tab:best hyperparameters} lists the best hyperparameters for four networks trained with different training and validation dataset sizes. The optimal hyperparameters do not differ significantly among these networks. We compared the performance of different models by calculating the MSE on a normalized test dataset of 200 samples. Additionally, we use the coefficient of determination $R^2$ as a metric of how well the model predicts given outcomes. The $R^2$ score can take values up to 1, which corresponds to an exact match of predicted and true values. For the four NN models the match is very good with an $R^2$ score higher than 0.99. Models with more training data perform better overall with only small improvements for large datasets.

\begin{table}[ht]
\centering
\footnotesize
\begin{tabular}{|l|llll|}
\hline
Number of samples (training + validation) & 250 & 500 & 750 & 1800 \\ \hline
Hidden layers                                                      & 1                                & 1                                & 1                                & 1                                 \\
Hidden dimension                                                   & 682                              & 827                              & 676                              & 682                               \\
Learning rate [$\times 10^{-4}$]                                                     & 2.34                         & 1.31                         & 1.33                         & 1.37                          \\
Batch size                                                         & 16                               & 8                                & 8                                & 16                                \\ \hline
MSE on test dataset (200 samples) [$\times 10^{-5}$]                                      & 182.66                      & 52.78                       & 22.17                       & 4.11                        \\
$R^2$ on test dataset (200 samples)                                       & 0.9982                      & 0.9995                       & 0.9998                      & 0.9999                       \\ \hline

\end{tabular}
\caption{Best hyperparameters and performance for neural networks with different number of samples (training + validation). The normalized test dataset consists of 200 samples not included in any other datasets.}
\label{tab:best hyperparameters}
\end{table}

\begin{figure}[h]
    \centering
    \includegraphics[width=0.8\textwidth]{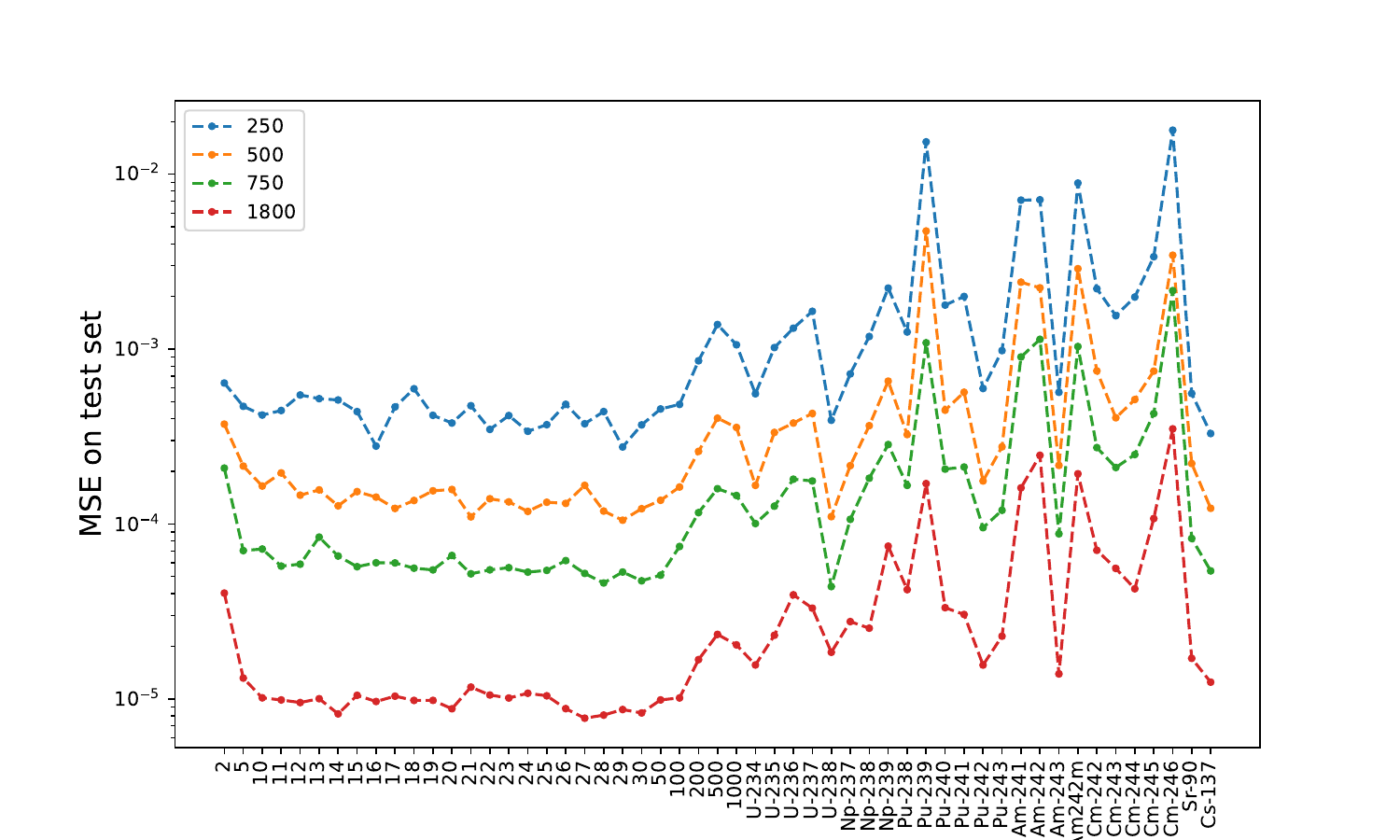}
    \caption{MSE on normalized test dataset (200 samples) of best models reported in Table \ref{tab:best hyperparameters}. The single numbers indicate the DH after $x$ years.}
    \label{fig:individual mse}
\end{figure}

Figure \ref{fig:individual mse} shows the MSE on the test dataset of every output for the best networks reported in table \ref{tab:best hyperparameters}. The error of the decay heat is generally lower than most of the nuclide concentrations. Some of the nuclide outputs, such as $^{242}$Am and $^{246}$Cm, exhibit a much larger error, independent of the model used.

As the generation of training data relies on computationally expensive CASMO5 simulation we must choose the model carefully to find the balance between accuracy and computational cost. For further evaluations we chose the best model obtained with 500 samples.

\begin{figure}[h!]
    \centering
    \includegraphics[width=\textwidth]{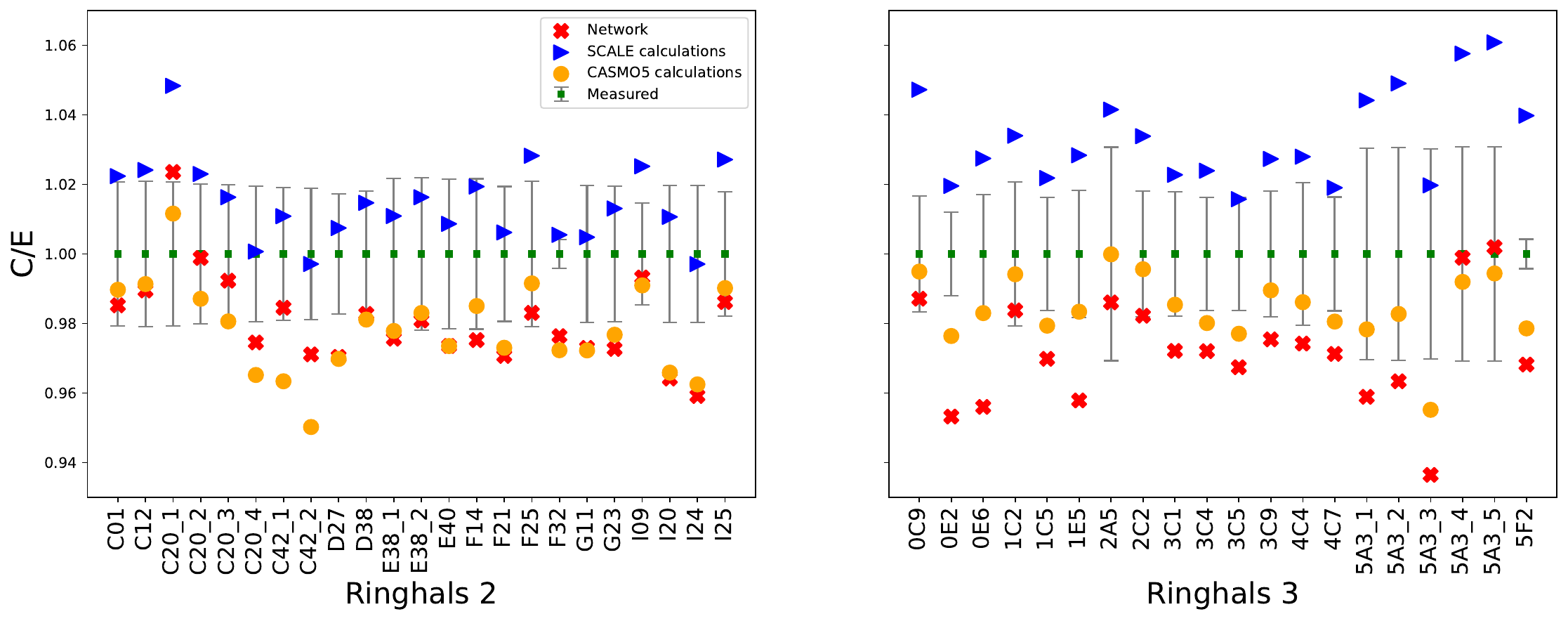}
    \caption{Decay heat measurements and network prediction with uncertainties as 2 standard deviations. Measurements and SCALE calculations obtained from \cite{sturek_measurements_2006}.}
    \label{fig:Fuel assembly measurements}
\end{figure}

To compare the network with experimental data we present the ratio $C/E$ between calculated DH $C$ and measurements $E$. Figure \ref{fig:Fuel assembly measurements} shows $C/E$ of all 43 measurements from the Clab facility for the NN and CASMO5. The SCALE calculations from the Clab reported were also included for reference. Table \ref{tab:DH bias} reports the average bias of the calculations. The network predicts the DH with a similar accuracy to the computational models.
As with all computational tools, the reported bias has to be taken into account when using the network for safety assessment of SNF management.

\begin{table}[h!]
    \centering
    \begin{tabular}{|lll|}
        \hline
        & Ringhals 2 & Ringhals 3 \\
        SCALE   & +1.47 \%  & +3.31 \%  \\
        CASMO5   & -2.15 \%  & -1.57 \%  \\
        Network & -1.72 \%  & -2.31 \%  \\ \hline
    \end{tabular}
    \caption{Average bias of decay heat prediction from different models with respect to experimental measurements from Clab report \cite{sturek_measurements_2006}.}
    \label{tab:DH bias}
\end{table}

\subsection{Speedup}

The computational speedup could be calculated as $\frac{T_C}{T_{evaluation}}$ with $T_C$ the time required for a CASMO5 simulation and $T_{evaluation}$ that of an NN evaluation, which would correspond to a speedup of approximately $10^5$. However, this calculation is biased as it does not take into account the computational cost of training the network and generating the training samples. Therefore, we define the speedup $S$ gained with the NN compared to CASMO5 as a function of the number of samples to evaluate $N$ as 
\begin{equation}\label{eq:speedup}
    S(N) = \frac{N\cdot T_C}{T_{Network}(N)}
\end{equation}
where $T_C = 58 \pm 11$ seconds is the CASMO5 simulation time and the network time $T_{Network}$ is defined as

\begin{equation*}\label{eq:NN time}
    T_{Network}(N) = T_{train} + N\cdot T_{evaluation} + N_{train}\cdot T_C\,,
\end{equation*}
where $T_{train} = 105 \pm 0.6$ seconds is the time required to train the network, $T_{evaluation} = (5\pm0.4)\cdot10^{-4}$ seconds is the time required for a forward prediction with the NN, and $N_{train}=500$ is the number of required training samples with CASMO5. When evaluating a large number of samples, or when the CASMO5 simulation time is large, the speedup behaves like $N/N_{train}$ as the network evaluation and training time become negligible. Note, that we exclude the time needed to find optimal hyperparameters when discussing the speedup. The reported uncertainties are statistical variations obtained by repeated measurements. As an example in this work, where two UQ and SAs were carried out, i.e. $N=5072$ in (eq. \ref{eq:speedup}), the total speedup was of more than 10.





\subsection{Uncertainty Quantification and Sensitivity Analysis}
\label{sec:UQandSA}

\begin{figure}[h!]
    \centering
    \begin{subfigure}[b]{0.32\textwidth}
        \includegraphics[width=\textwidth]{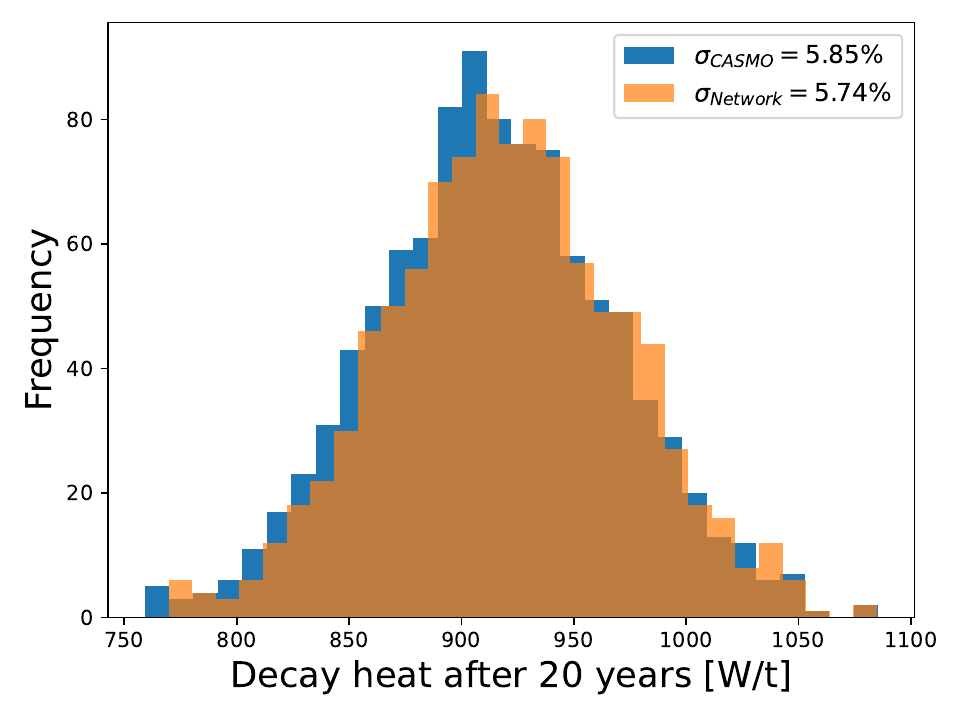}
    \end{subfigure}
    \hfill
    \begin{subfigure}[b]{0.32\textwidth}
        \includegraphics[width=\textwidth]{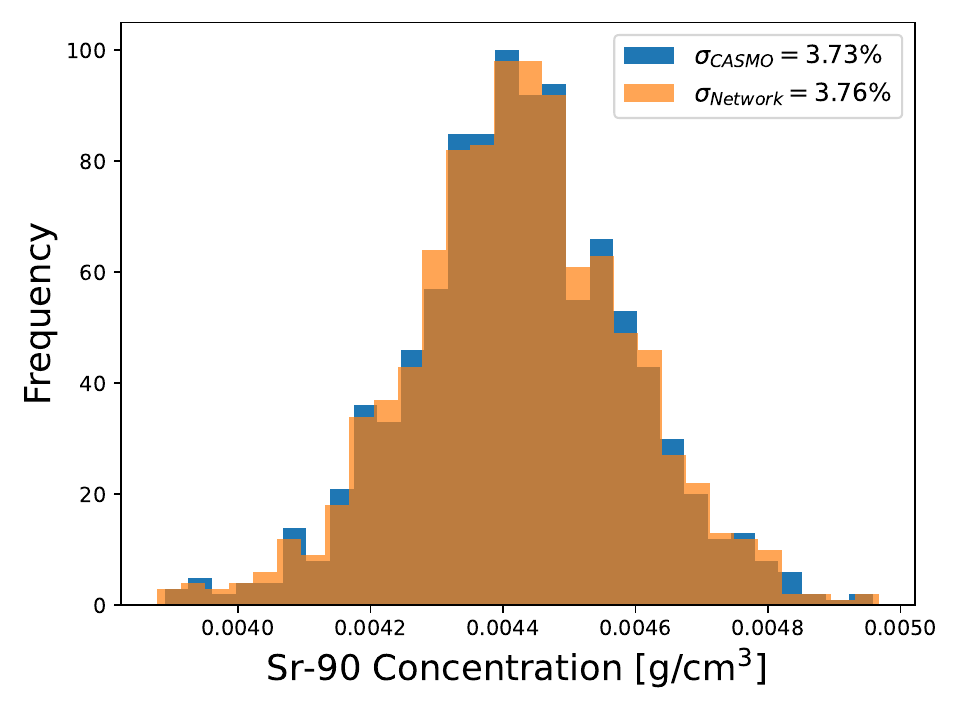}
    \end{subfigure}
    \hfill
    \begin{subfigure}[b]{0.32\textwidth}
        \includegraphics[width=\textwidth]{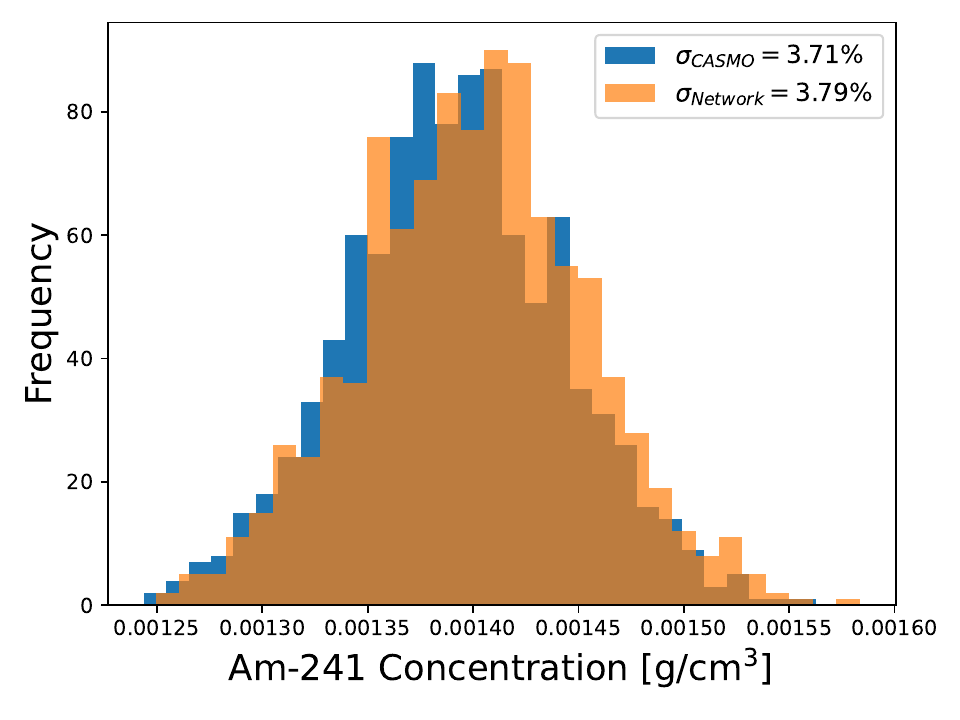}
    \end{subfigure}
    \caption{Distribution and relative standard deviation of three outputs over 1000 samples around FA C20. CASMO5 simulation in blue, network prediction in orange.}
    \label{fig:UQ histograms C20}
\end{figure}

\begin{figure}[h!]
    \centering
    \begin{subfigure}[b]{0.32\textwidth}
        \includegraphics[width=\textwidth]{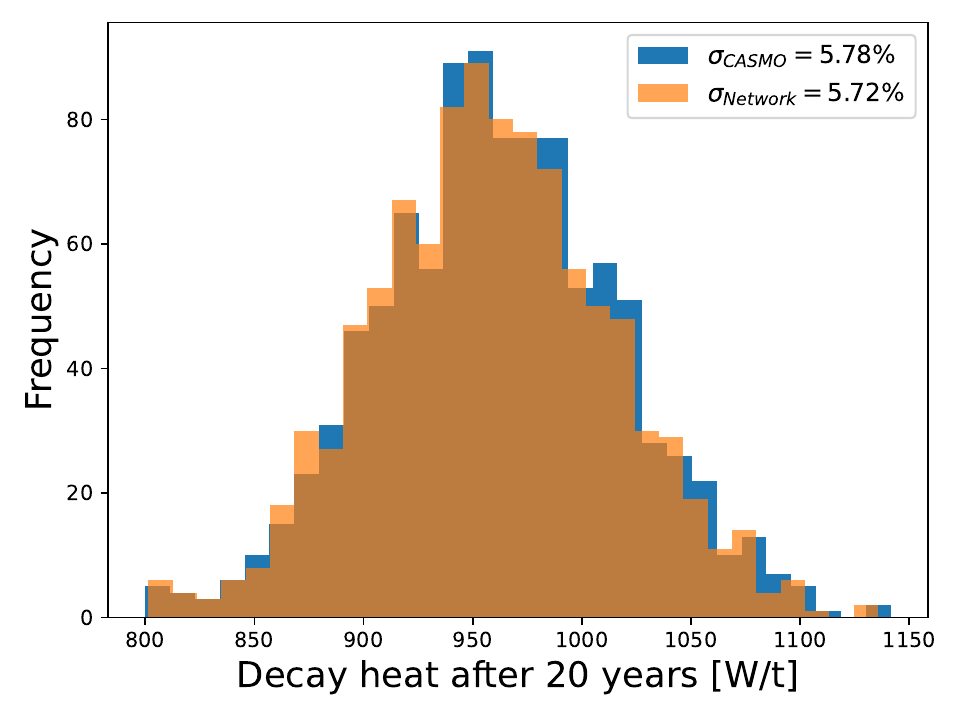}
    \end{subfigure}
    \hfill
    \begin{subfigure}[b]{0.32\textwidth}
        \includegraphics[width=\textwidth]{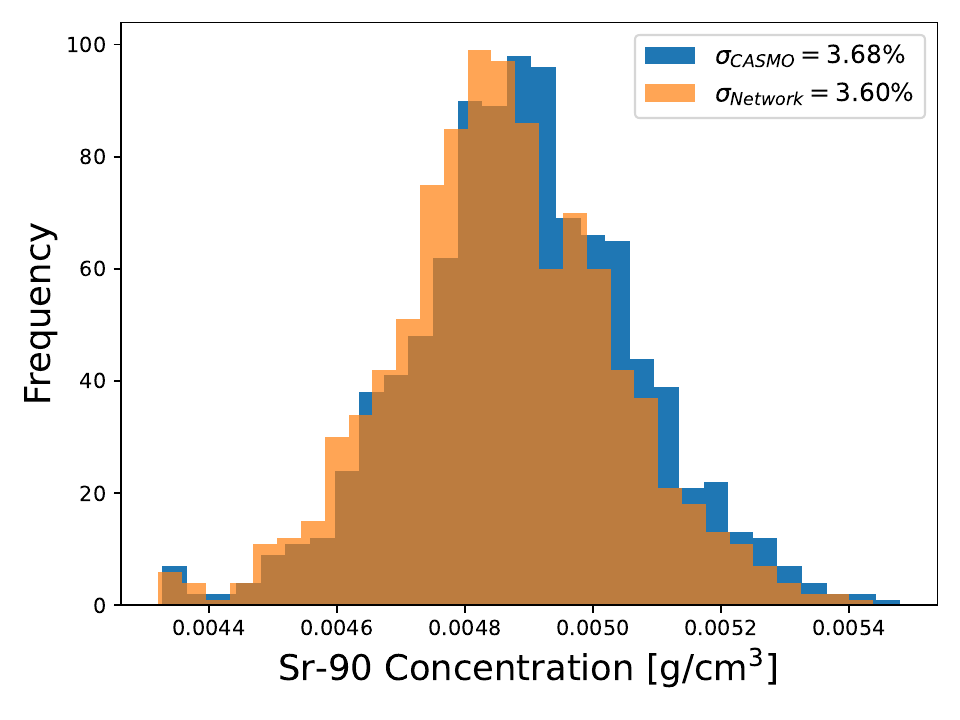}
    \end{subfigure}
    \hfill
    \begin{subfigure}[b]{0.32\textwidth}
        \includegraphics[width=\textwidth]{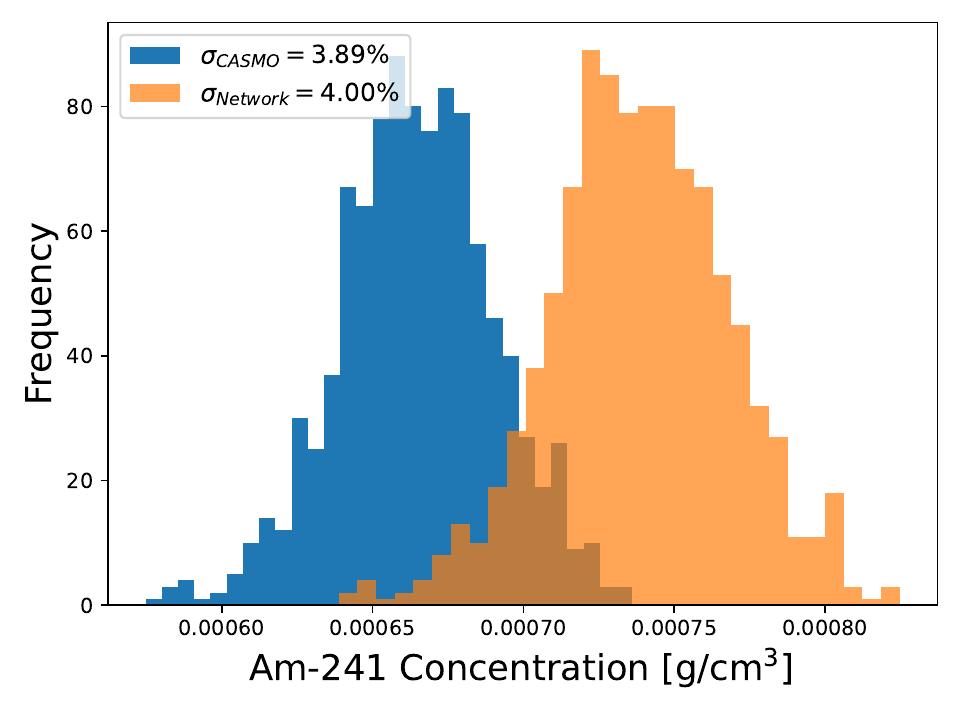}
    \end{subfigure}
    \caption{Distribution and relative standard deviation of three outputs over 1000 samples around FA C01. CASMO5 simulation in blue, network prediction in orange.}
    \label{fig:UQ histograms C01}
\end{figure}

For UQ, we compared the network predicted and CASMO5 calculated distribution of outputs. Therefore, 1000 samples for FA C20 were taken with an input uncertainty of 5\%. Figure \ref{fig:UQ histograms C20} shows the DH after 20 years and the concentration of $^{90}$Sr and $^{241}$Am for both NN and CASMO5 simulations. The mean and standard deviation of the network predictions are consistent with the predictions of CASMO5. This was the FA the network was trained for. In addition the UQ was performed for FA C01. Figure \ref{fig:UQ histograms C01} shows the distributions of the same outputs around FA C01. The DH and $^{90}$Sr concentration agree well with CASMO5 simulations. 
However, the predicted and simulated distributions of $^{241}$Am concentration differ largely. The mean of the network predicted distribution is larger by about 11\%, whereas the standard deviation is similar and also the shape of the distribution. Looking at figure \ref{fig:individual mse} we would expect less accurate predictions for $^{241}$Am compared to DH after 20 years and $^{90}$Sr. Considering that the training dataset for the network was based on FA C20, it is also reasonable to assume that the network performs worse around other FAs. Nonetheless, the predicted relative standard deviation $\sigma/\mu$ agrees with the calculated one.

\begin{figure}[h]
    \centering
    \includegraphics[width=0.6\textwidth]{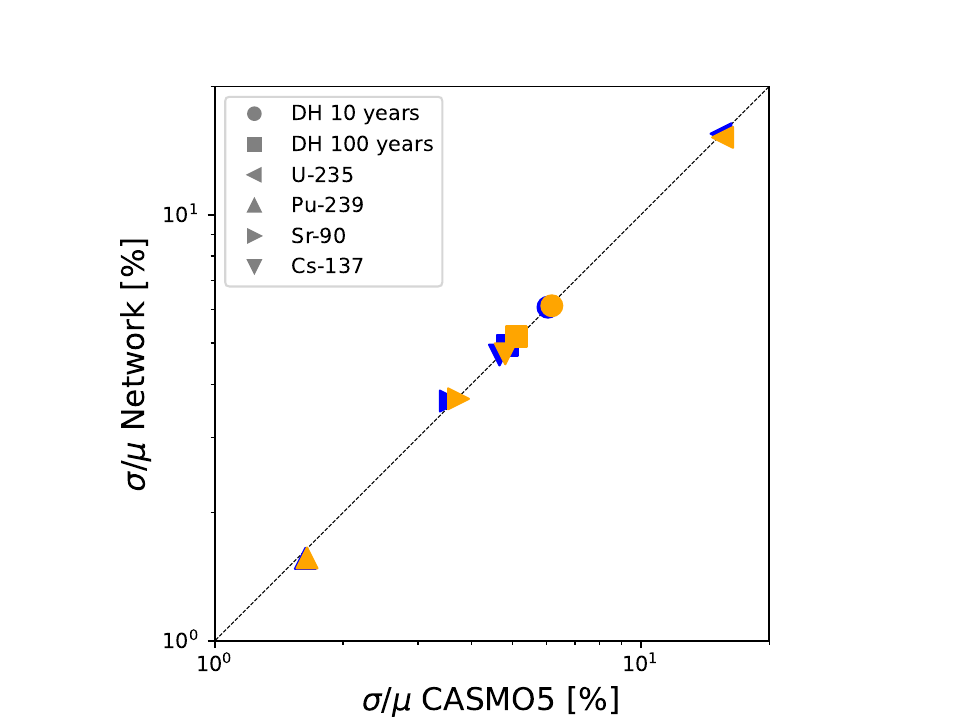}
    \caption{Predicted relative standard deviation against CASMO5 for FA C01 (blue) and C20 (orange). Statistical variations obtained from bootstrapping 10000 samples are too small to be visible.}
    \label{fig:compare standard deviation}
\end{figure}

Figure \ref{fig:compare standard deviation} shows the NN predicted relative standard deviation $\sigma/\mu$ against CASMO5 simulations for several outputs from FAs C01 and C20. The plotted values were obtained via bootstrapping by resampling 1000 samples 10000 times and measuring the mean and variance of the standard deviation. We observe, that most predicted values coincide with the calculated ones for both FAs.
Although the accuracy of the predicted mean value depends on the MSE of the outputs, we can see that the predicted relative standard deviation is comparable to CASMO5 simulations.

\begin{figure}[h!]
    \centering
    \begin{subfigure}[b]{0.475\textwidth}
        \centering
        \includegraphics[width=\textwidth]{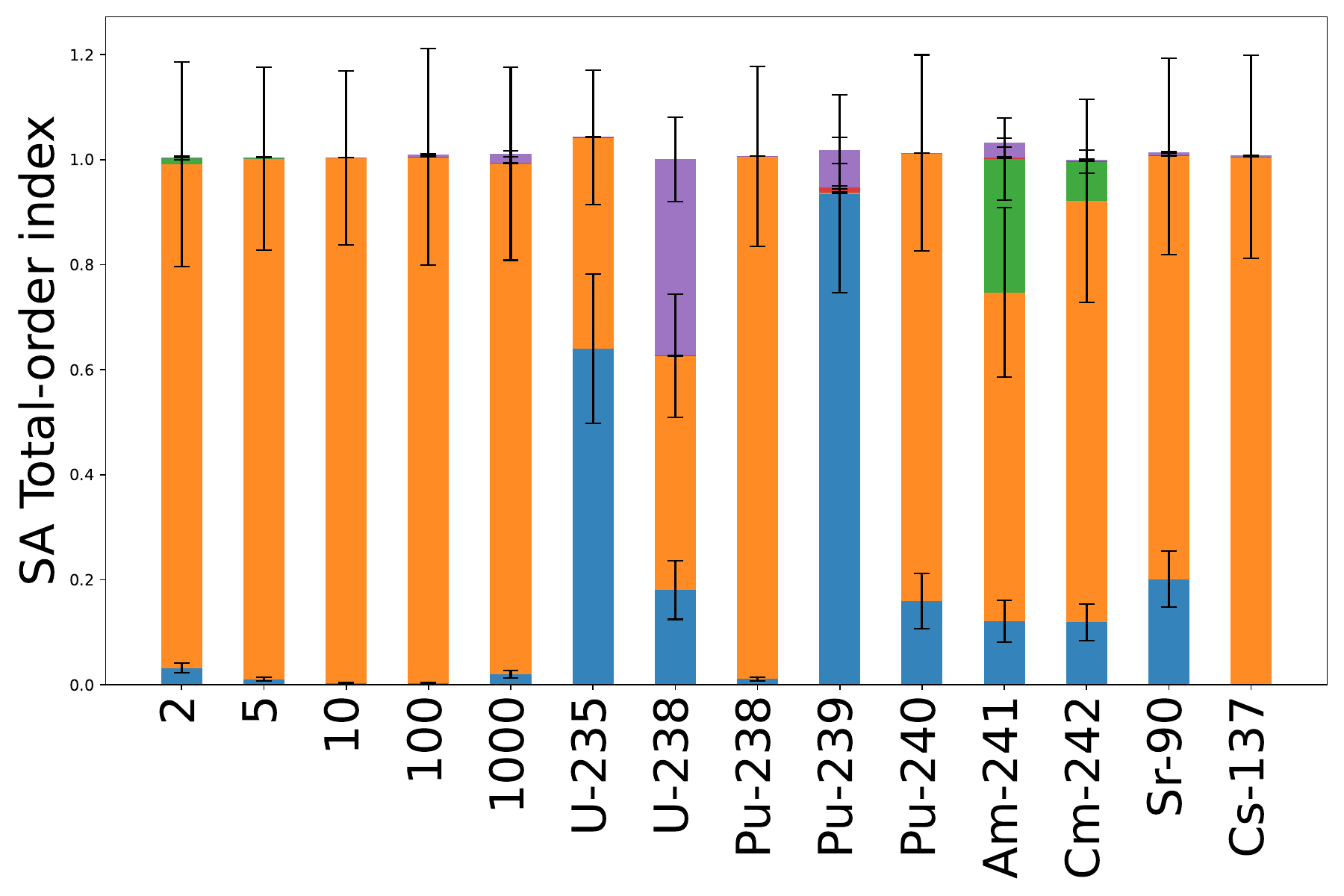}
        \caption{C01}
        \label{fig:SA C01 Network}
    \end{subfigure}
    \hfill
    \begin{subfigure}[b]{0.475\textwidth}
        \centering
        \includegraphics[width=\textwidth]{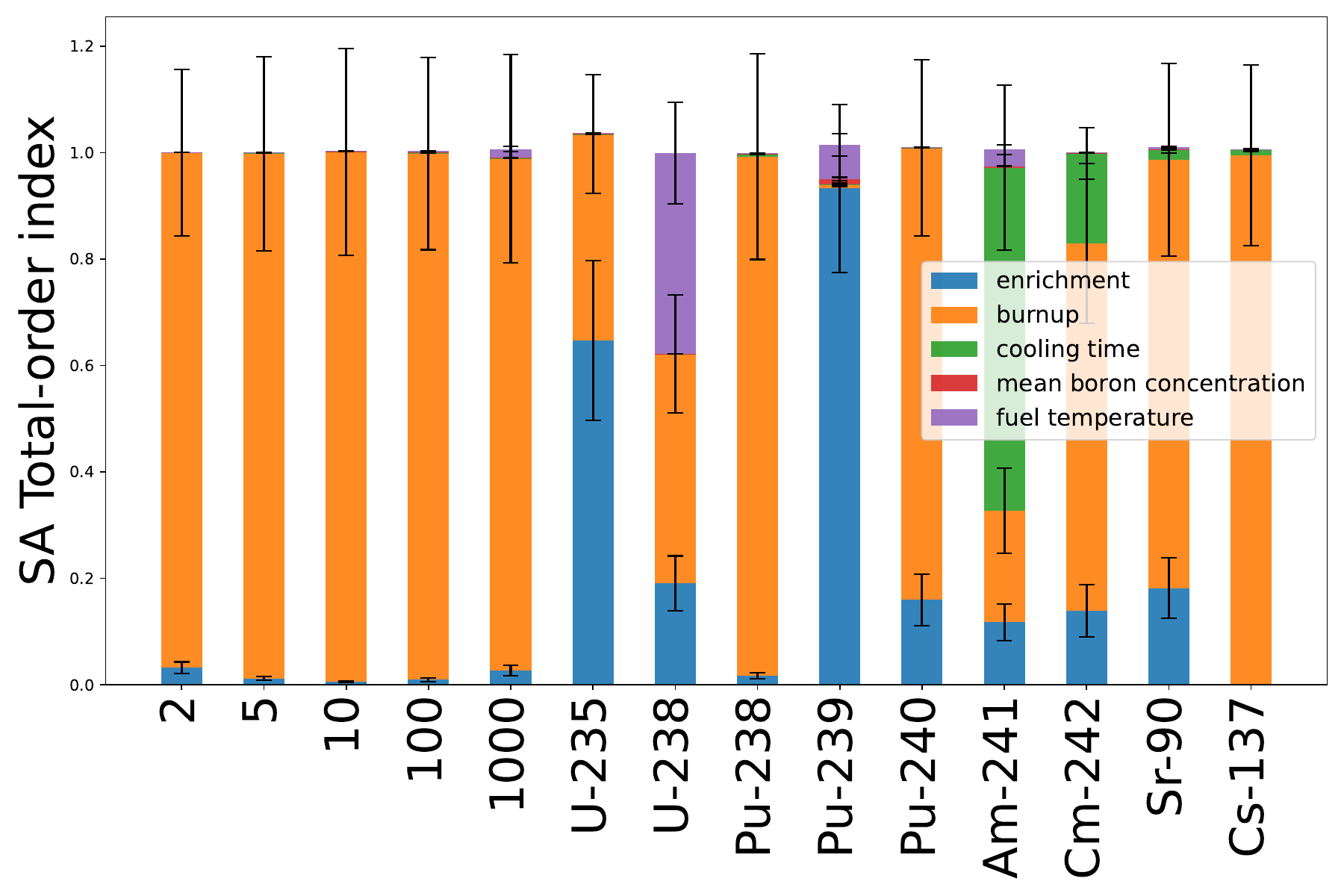}
        \caption{C20}
        \label{fig:SA C20 Network}
    \end{subfigure}
    \caption{Total-order Sobol' index based on the network prediction of 1536 samples with 5\% uncertainty around FAs C01 and C20. The single numbers indicate the DH after 2, 5, 10, 100 and 1000 years.}
    \label{fig:SA total order index}
\end{figure}

In addition to UQ we were also interested in finding the most influencing input parameters for each output. Hence a sensitivity analysis was conducted. 
Figure \ref{fig:SA total order index} shows the total-order Sobol' SA index based on 1536 samples evaluated with the network for FAs C01 and C20. Specifically, shown is the importance of every input to some output quantities, such as the DH after 100 years and the isotopic concentration of $^{137}$Cs. For both analyzed FAs, the Sobol' method returns practically identical results between samples evaluated by the network and CASMO5. When analyzing the DH, we observe that the burnup is considerably more important than all other inputs. This implies that to best reduce the uncertainty of DH one should aim to reduce the uncertainty of the burnup. When devising safety measures for nuclear waste management, a small uncertainty in DH is crucial as this comes with a reduction of SNF canisters and associated costs \cite{Rochman2023,SOLANS2020110897}. Furthermore, we observe that the cooling time has a bigger impact on the concentration of $^{241}$Am for C20 fuel assembly.

\section{Conclusion}

In this paper we have presented a surrogate modeling approach for the characterization of spent nuclear fuel using neural networks. By training the NN with data generated from CASMO5 lattice calculations, we achieved accurate predictions of decay heat and nuclide concentrations of SNF as a function of key input parameters, including enrichment, burnup, cooling time, mean boron concentration and fuel temperature. The training dataset consisted of 500 CASMO5 simulations, and once trained, the NN could be evaluated in less than a millisecond. Although the NN achieved a mean squared error of less than $10^{-2}$ for all predicted quantities, it performed notably better at predicting the decay heats than nuclide concentrations. For certain nuclides the prediction error was an order of magnitude larger than for the decay heat. For future work it could be of interest to investigate these differences, and consider training separate networks for the decay heat and nuclide concentrations.

The NN predictions demonstrated a deviation within 3\% from experimentally measured decay heat, indicating a similar level of accuracy to CASMO5 and SCALE computational models. Importantly, this accuracy was consistent across different fuel assemblies of the same or similar architecture. 

Moreover, the NN was utilized for uncertainty quantification and sensitivity analysis, exhibiting comparable results to CASMO5, while significantly reducing computational time. A speedup of a factor 10 was achieved for the uncertainty quantification and sensitivity analysis of two assemblies. Note however that the computational costs would be reduced even further if UQ and SA were carried out for more assemblies. This remarkable computational efficiency greatly accelerates the analysis process compared to traditional physics-based models without compromising accuracy. 
Hence, the findings presented in this paper suggest that the NN approach has the potential to enhance efficiency in assessing nuclear fuel behaviour and associated risks.
However, further investigations should explore the capability of the NN approach to a broader range of SNF types, such as mixed oxide fuel or boiling water reactor fuel.

To further improve the accuracy of the predictions with respect to measurements, one possible direction would consist in incorporating available measurements of SNF, like the ones used for model validation in this work, in the training dataset, in addition to physics-based models. This could improve prediction accuracy and potentially reduce the number of CASMO5 simulations required for training.
Another approach for increasing the accuracy could be to use several different codes for generating the training dataset in order to get rid of the bias introduced by CASMO5. 

Within this work the uncertainty quantification was a proof-of-concept, assuming that the uncertainty of each quantity is 5\% to demonstrate the capabilities and speedup of the NN. To address real-world applications, it is important to consider the real uncertainties associated with input parameters during uncertainty quantification. 

Overall, this study underscores the potential of NNs as surrogate models in advancing the field of SNF characterization and contributing to the safety and efficiency of nuclear waste management.

\section{Acknowledgements}
This work was part of the COLOSS and COLOSS-2 projects (COmbined Loading Optimization with
Simulations and Surrogate models), funded by Swissnuclear.

\bibliographystyle{elsarticle-num} 
\bibliography{ref}

\appendix

\section{Generation of Input Files for Training Data}
\label{app:generation_input}

The simulations for the training dataset were all done with a modified version of the input file for the C20 assembly. This FA was one of the 34 PWR assemblies presented in the Clab report \cite{sturek_measurements_2006}. It was a $15\times15$ UO2 FA (204 fuel rods, 21 water rods), and was burned in the PWR Ringhals 2 during 4 cycles. It had an enrichment of $3.095\%$, and an average fuel temperature of 887 K. Its fuel history is shown in table \ref{tab:history_C20}, from which the burnup can be calculated to be 35.72 MWd/kgU, the number of cooling days 2068, and the average boron concentration 310.75 ppm.

\begin{table}[h!]
    \centering
    \renewcommand{\arraystretch}{1.2} 
    \def\colwidth{2.4cm}
    \begin{tabular}{|p{\colwidth}|p{\colwidth}|p{\colwidth}|p{\colwidth}|p{\colwidth}|}
    \hline
        Cycle Type & Duration [days] & Burnup [MWd/kgU] & Boron Concentration [ppm] & Power [W/gU]\\
        \hline
        Burnup & - & 11.247 & 143 & 10.9\\
        Cooling  & 85 & - & - & -\\
        Burnup & - & 9.377 & 459 & 35.1\\
        Cooling & 56  & - & - & -\\
        Burnup & - & 7.454 & 342 & 23.9 \\
        Cooling & 1927 & - & - & -\\
        Burnup & - & 7.642 & 299 & 28.7\\
        \hline
         & Total=2068 & Total=35.72 & Mean=310.72 & \\
         \hline
    \end{tabular}
    \caption{Reported burnup history of the C20 assembly in the Clab report. Note that in the case of burnup cycles, the duration is ommitted as it would be redundant, since it can be calculated with $\frac{burnup}{power}$.}
    \label{tab:history_C20}
\end{table}

To generate different input files for the training dataset, only the five input quantities of interest (tab. \ref{tab:input_ranges}) were modified from the C20 input file. These five input quantities, mentioned in section \ref{sec:prep_training_set}, were enrichment, fuel temperature, mean boron concentration, cooling time between cycles, and burnup. The former two quantities are given by a single value, and could be trivially changed in the input file. However, the three latter quantities are calculated as a sum of terms from the burnup history as in table \ref{tab:history_C20}, and could not be modified with a single change of number in the input file. The approach taken was to multiply the boron concentration, cooling time, and burnup in each cycle by a factor, such that the overall average quantities would be modified for each simulation, but the ratio between the cycles was kept constant. 

To clearly illustrate the procedure, let the randomly sampled quantities be \texttt{enrichment}, \texttt{fuelTemp}, \texttt{bor}, \texttt{coolTime}, and \texttt{burnup}. Then, the generated CASMO5 input file would have an enrichment \texttt{enrichment}, a fuel temperature \texttt{fuelTemp} for all cycles, and a burnup history as shown in table \ref{tab:modified_hist}.

\begin{table}[h!]
    \centering
    \renewcommand{\arraystretch}{1.5} 
    \def\colwidth{2.6cm}
    \begin{tabular}{|p{\colwidth}|p{\colwidth}|p{\colwidth}|p{\colwidth}|p{\colwidth}|}
    \hline
        Cycle Type & Duration [days] & Burnup [MWd/kgU] & Boron Concentration [ppm] & Power [W/gU]\\
        \hline
        Burnup & - & $\frac{11.247}{35.72}\cdot\texttt{burnup}$ & $\frac{143}{310.75}\cdot\texttt{bor}$ & 10.9\\
        Cooling  & $\frac{85}{2068}\cdot\texttt{coolTime}$ & - & - & -\\
        Burnup & - & $\frac{9.377}{35.72}\cdot\texttt{burnup}$ & $\frac{459}{310.75}\cdot\texttt{bor}$ & 35.1\\
        Cooling & $\frac{56}{2068}\cdot\texttt{coolTime}$ & - & - & -\\
        Burnup & - & $\frac{7.454}{35.72}\cdot\texttt{burnup}$ & $\frac{342}{310.75}\cdot\texttt{bor}$ & 23.1\\
        Cooling & $\frac{1927}{2068}\cdot\texttt{coolTime}$ & - & - & -\\
        Burnup & - & $\frac{7.642}{35.72}\cdot\texttt{burnup}$ & $\frac{299}{310.75}\cdot\texttt{bor}$ & 28.7\\
        \hline
    \end{tabular}
    \caption{Burnup history in a given sample of the training dataset. The duration of the burnup cycles is omitted since it is calculated from the power and burnup.}
    \label{tab:modified_hist}
\end{table}

\end{document}